\documentclass[journal]{./IEEEtran}
\usepackage{cite}
\ifCLASSINFOpdf
	\usepackage[pdftex]{graphicx}
	\graphicspath{{./}}
\else
	\usepackage[dvips]{graphicx}
	\graphicspath{{./}}
\fi
\usepackage{epstopdf}
\usepackage{amsmath}
\usepackage{amssymb}
\usepackage{algorithmicx}
\usepackage{algpseudocode}
\usepackage{algorithm}
\usepackage{array}
\usepackage[caption=false,font=footnotesize]{subfig}

\usepackage{url}
\begin{document}
\title{ADMM-based Networked\\Stochastic Variational Inference}
\author{Hamza~Anwar,~\IEEEmembership{Member,~IEEE,}
        and~Quanyan~Zhu,~\IEEEmembership{Member,~IEEE}
\thanks{This work is partially supported by the grants EFMA-1441140 and SES-1541164 from National Science Foundation.}~\thanks{H. Anwar and Q. Zhu are with the Department of Electrical and Computer Engineering,
New York University, New York, NY 10003 USA (e-mail: ha1082@nyu.edu, qz494@nyu.edu).}
\thanks{Manuscript received , 2017; revised \today.}}
\markboth{*}%
{\ifCLASSOPTIONpeerreview Anwar \MakeLowercase{\textit{et al.}}: ADMM-based Networked Stochastic Variational Inference}
\maketitle
\begin{abstract}
Owing to the recent advances in ``Big Data'' modeling and prediction tasks, variational Bayesian estimation has gained popularity due to their ability to provide exact solutions to approximate posteriors. One key technique for approximate inference is stochastic variational inference (SVI) \cite{hoffman2013stochastic}. SVI poses variational inference as a stochastic optimization problem and solves it iteratively using noisy gradient estimates. It aims to handle massive data for predictive and classification tasks by applying complex Bayesian models that have observed as well as latent variables. This paper aims to decentralize it allowing parallel computation, secure learning and robustness benefits. We use Alternating Direction Method of Multipliers in a top-down setting to develop a distributed SVI algorithm such that independent learners running inference algorithms only require sharing the estimated model parameters instead of their private datasets. Our work extends the distributed SVI-ADMM algorithm that we first propose, to an ADMM-based networked SVI algorithm in which not only are the learners working distributively but they share information according to rules of a graph by which they form a network. This kind of work lies under the umbrella of `deep learning over networks' and we verify our algorithm for a topic-modeling problem for corpus of \emph{Wikipedia} articles. We illustrate the results on latent Dirichlet allocation (LDA) topic model in large document classification, compare performance with the centralized algorithm, and use numerical experiments to corroborate the analytical results.
\end{abstract}
\IEEEpeerreviewmaketitle
\section{Introduction}
\IEEEPARstart{T}{he} explosive influx of data and information for modern day technological systems has opened doors to revolutionary possibilities. One of the most vital uses of this data is in modeling, visualizing, and analyzing large data sets through probabilistic tools. Statistical machine learning is at the core of numerous such applications in what is becoming known as Internet of Things (IoT). Such iterative learning mechanisms help in better control performance for many cyber-physical-systems in which estimation of parameters and system-identification is required. Probabilistic graphical modeling is one key research area that has helped in data analysis in inference and prediction tasks, \cite{koller2009probabilistic}. These models visually express assumptions about data and its hidden structure. Posterior inference algorithms have been proven to exploit such models in explaining this hidden structure while being adaptive, robust, parallelizable, and scalable.

\emph{Variational inference}, from late 90s, is a method that transforms complex inference problems into high dimensional optimization problems. In contrast to Monte-Carlo sampling methods (that simply aim to find exact answer to an approximate problem), the variational Bayesian approach solves for optimal solution under constraints to the right inference problem, \cite{jordan1999introduction}. On the same lines, stochastic variational inference (SVI) was developed recently that extends variational inference to be solved using stochastic optimization under certain assumptions, \cite{hoffman2013stochastic}. SVI works iteratively in gradient ascent fashion using noisy gradient estimates. It provides approximate model posteriors with only a few passes through a large data collection, making it highly scalable. We propose ADMM-based Networked SVI -- a distributed stochastic variational inference technique that builds upon standard SVI, retaining most of its benefits, based on the highly parallelizable alternating direction method of multipliers (ADMM) where the agents are connected as in a graph with nodes and edges.

\subsection{Related Work}Numerous extensions have been proposed for the SVI framework in its application to more model classes (by \cite{foti2014stochastic} and \cite{johnson2014stochastic}), different underlying processes (by \cite{hensman2013gaussian} and \cite{gal2014distributed}), and structural exploitations (by \cite{hoffman2015structured}) making it faster and widely deployable. A variety of works focus on making variational methods distributed to enhance parallelizability. The work on distributed Bayesian nonparametric models by \cite{campbell2015streaming} is commendable in making variational inference updates distributed, asynchronous, and `streaming' (online) and they've shown it to outperform standard SVI. However, their work is only specific to the Dirichlet process mixture and lacks in generalizability to the class of probabilistic models that SVI can deal with. Similar to \cite{campbell2015streaming}, another work D-MFVI by \cite{babagholami2015d} uses ADMM for decentralizing, like us, but lacks in being extendable to online updates, fast convergence rate, and other desirable properties of SVI. Distributed VBA (\cite{hua2016distributed}) also uses ADMM however their approach lacks in scalability to large data without demanding adequate computational resources.

None of these works use stochastic optimization methods to speed up inference and hence they are fundamentally different from standard SVI itself. One recent work `Extended-SVI' by \cite{raman2016extreme} retains the benefits of SVI while making it distributed and asynchronous. They employ a rather simple algorithmic change to SVI however their work remains unexplored in terms of depth because they particularly focus on Gaussian mixture models and do not provide how it is extendable to all other probabilistic models for which SVI in general works.

In contrast to all related works highlighted, our approach extends the general SVI framework to a networked stochastic optimization consensus problem. We tackle the issue of generalizability to all graphical models by providing a general solution and make use of the stochastic gradient updates that make it fast. We run ADMM updates along with stochastic gradient ascent for variational objective to reach consensus among a number of distributed learners. SVI itself being a non-convex stochastic optimization problem makes the distributed problem trickier. Our work aims to show that independent learners that use SVI for similar applications, can collaborate by exchanging their results (not the data itself) to benefit from each other improving overall accuracy of results. This approach makes our work unique and applicable to wider distributed large-scale inference problems. Not just that, but this kind of an approach poses a game problem with multiple agents interested in performing their own inference tasks, and simultaneously benefiting from each other through reinforcement learning and cooperation.

\section{ADMM-based Distributed SVI}
Building upon the recent work on SVI by Hoffman et al., \cite{hoffman2013stochastic}, we consider the SVI problem for a network of learners.

The $N$ observations are $x = x_{1:N}$; the vector of global hidden variables is $\beta$; the $N$ local hidden variables are $z = z_{1:N}$, each of which is a collection of $J$ variables $z_n = z_{n,1:J}$; the vector of fixed parameters is $\alpha$. (Note we can easily allow $\alpha$ to partly govern any of the random variables, such as fixed parts of the conditional distribution of observations. To keep notation simple, we assume that they only govern the global hidden variables.)
\subsection{Optimization problem}
\begin{align*}
\min_{\lambda_k} ~~~ &\sum_{k=1}^K g_k(\lambda_k)\\
\text{subject to} ~~~ &\lambda_k - \zeta = 0,\quad k= 1,\cdots, K
\\&\lambda_k \in \Gamma_k
\end{align*}
where each $\lambda_k$ is an $m$-sized vector and $\Gamma_k$ indicates the feasible set for the variables $\lambda_k$ (typically $\Gamma_k = \mathbb{R}^m_+$) and,
\begin{align}
\label{eq:cent_obj_func}
\begin{split}
g_k(\lambda_k) := -\mathbb{E}_{\phi(\lambda_k)}[\eta_g(x,z)] ^{\top} \nabla_{\lambda_k} a_g(\lambda_k) + \lambda_k^{\top}\nabla_{\lambda_k} a_g(\lambda_k)\\ - a_g(\lambda_k) + \text{const.}
\end{split}
\end{align}
which is the standard SVI problem objective function for a single learner. The above optimization problem gives us a solution for $K$ learners when they form a consensus. Using an Augmented Lagrangian approach, as in ADMM, we solve this problem in a distributed iterative fashion for multiple learners.
\subsection{ADMM-based solution}\label{sec:dist_svi_solution}
Augmented Lagrangian with a quadratic penalty is used to arrive at the ADMM update iterations. The Lagrange multipliers are denoted by $y_k\in \Gamma_k$. Minimization updates for each processor/agent are given as:
\begin{align*}
\label{eq:admm_it}{\lambda}_k^{t+1} ~=&~ \arg\min_{\lambda_k} \left(g_k(\lambda_k) + (\lambda_k - \zeta^t)^\top y_k^t + \frac{c}{2}||\lambda_k-\zeta^t||_2^2 \right),\\
\zeta^{t+1} ~:=&~ \frac{1}{K}\sum_{k=1}^K \left( \lambda_k^{t+1} + (1/c)y_k^t \right),\\
y_k^{t+1} ~:=&~ y_k^t + c (\lambda_k^{t+1}-\zeta^{t+1}).
\end{align*}
where $\zeta$ is called the {\it central collector}, and $c$ is the quadratic penalty parameter in the augmented Lagrangian which is given as:
\[L_c(\{\lambda_k\},\{y_k\},\zeta) = \sum_{k=1}^K g_k(\lambda_k) + (\lambda_k - \zeta) ^\top y_k+ \frac{c}{2}||\lambda_k-\zeta||_2^2.\]

Here, we note that the $\lambda$-minimization update which is actually a solution to $\lambda_k^{t+1} = \arg \min_{\lambda_k\in \Gamma_k} L_c(\{\lambda_k\},\{y_k^t\},\zeta^t)$, requires solving a constrained non-convex optimization problem. We solve this in a gradient descent fashion in of itself, as the standard SVI problem was also solved, but the original solution requires inversion of a Hessian matrix. For that we take into account one-step earlier value of $\lambda_k^t$ --- for details about derivation and Hessian inversion approximation used, see Appendix. Thereby, our proposed iterative ADMM methodology runs along with a gradient-descent iterative update of variables which is completely summarized in algorithm~\ref{algo:svi-admm}.
\begin{algorithm}
\caption{ADMM-based distributed SVI for $K$ players}
\label{algo:svi-admm}
\begin{algorithmic}[1]
\State{Initialize $c,\lambda_1^{(0)},\lambda_2^{(0)},\dots,\lambda_K^{(0)}$}
\State{Schedule step-size $\rho_t$ routine}
\Repeat
\For{$k\in\mathcal{K}$}
\State{Sample separate data points for all learners}
 \State{Use $x_k$ to compute its local variational parameters,
 \[\phi = \mathbb{E}_{\lambda_k^{t}}[\eta_l(x_k^{(N)},z_{k}^{(N)})].\]}
\State{Apply ADMM $\lambda$-minimization-update by computing intermediate global parameters $\hat{\lambda}_k$ and natural gradient,
\[\hat{\lambda}_k = \mathbb{E}_{\phi}[\eta_g(x_k^{(N)},z_k^{(N)})],\]
\[\hat{\nabla}_{\lambda_k^t} L_c = (\lambda_k^t -\hat{\lambda}_k) - \nabla^{-2} a_g(\lambda_k^t)(y_k^t + c(\lambda_k^t - \zeta^t)).\]
}
\State{Update the global variational parameters using gradient {ascent},
\[\lambda^{t+1}_k = \lambda^{t}_k + \rho^t (-\hat{\nabla}_{\lambda_k^t} L_c).\]}
\EndFor
\State{Update the central collector\begin{align*}\zeta^{t+1} ~=&~ \frac{1}{K}\sum_{k=1}^K \lambda_k^{t+1} + (1/c)y_k^{t}.
\end{align*}}
\State{Update all the Lagrange multipliers\begin{align*}y_k^{t+1} ~=&~ y_k^t + c (\lambda_k^{t+1}-\zeta^{t+1}).
\end{align*}}
\Until{forever}
\end{algorithmic}
\end{algorithm}
\subsection{Experimental results}
\begin{figure}[!t]
\centering
\includegraphics[width=\linewidth]{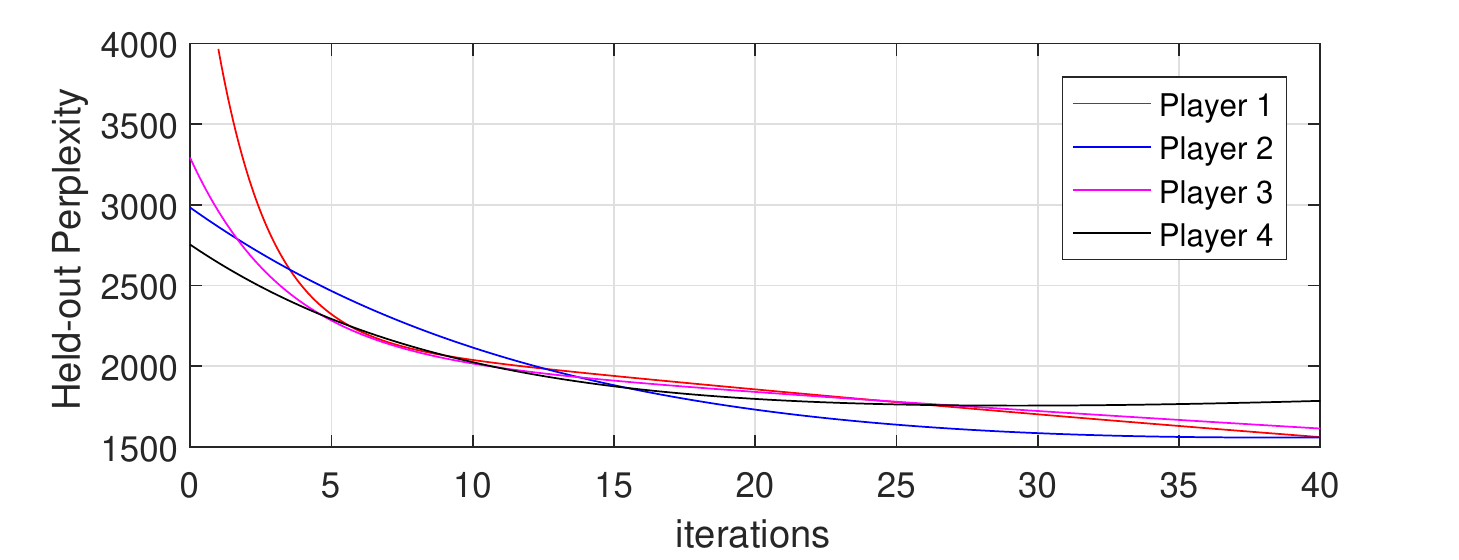}
\caption{Four players running independent learners and collaborating. This plot shows that the perplexity is decreasing over time.\label{fig:exp1}}
\end{figure}

\begin{figure}[!t]
\centering
\includegraphics[width=0.9\linewidth]{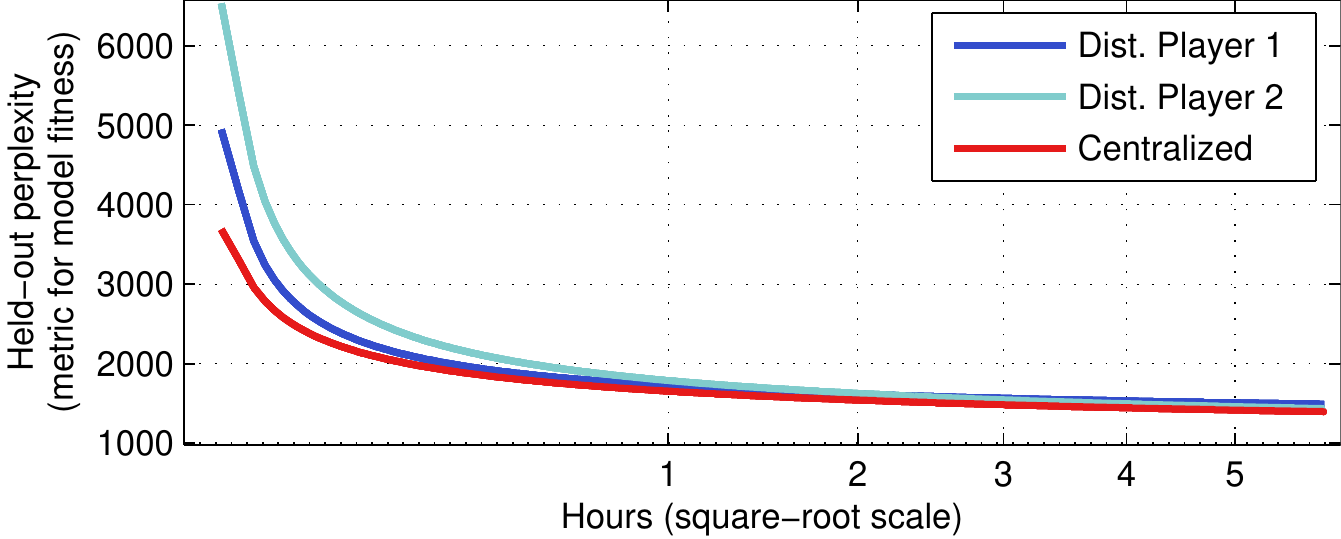}
\caption{Working for a two player network versus centralized algorithm.\label{fig:twoplayer}}
\end{figure}

New set of experiments for the distributed problem was performed with multiple learners. Here, we show the results with $4$ learners. Figure \ref{fig:exp1}, shows the convergence properties of our distributed learners, for a metric of the estimated model's fitness, known as the `held-out perplexity'. This same metric has been used by Hoffman et al. \cite{hoffman2010online} to show convergence of the algorithm. A comparison of centralized versus distributed two-player SVI algorithms is depicted in Figure \ref{fig:twoplayer}. We conclude that all the learners not only converge to higher precision in estimates (evident from the Figure \ref{fig:exp1}), but also achieve accuracy of estimates (evident from Table \ref{table:ADMM_four_players}), while simultaneously maintaining consensus.

From Table \ref{table:ADMM_four_players}, we see that the highly probable words for a given topic learned by any of the four learners all direct to a similar kind of subject. For example, in Topic\#98, the learners understood it to represent the names of months | even though for this topic, the distribution of word occurrences is different for the four players but it is evident that they all point to the same abstract class of words. Similarly, in other topics as well we see similarity in estimates. Sometimes, we even see that the descending order of the words is exactly the same, e.g., in Topic\#72 and Topic\#38 all learners have the same ordering of words. Thus, the table shows that despite the fact that all learners use their own independently fetched datasets from {\it Wikipedia} articles, the consensus between results is achieved among all the learners, due to the central collection constraint.

\begin{table}[!t]
\caption{Top three words for five topics learned by each of the four players after 35 iterations i.e. $64\times 35=2240$ independent documents analyzed by each player. Words are written in descending order of probability of occurrence. Penalty parameter for ADMM $c=5\times 10^{-8}$, and total topics were $100$.}
\label{table:ADMM_four_players}
\centering
\begin{tabular}{|c||c|c|c|c|}
\hline
&\textbf{Player 1}&\textbf{Player 2}&\textbf{Player 3}&\textbf{Player 4}\\
\hline\hline
&june& september&september&september \\
Topic\#98 &march&october&october& october \\ 
&november& november&november&november \\
\hline
&elected  &elected&elected &elected\\
Topic\#72 & democratic& democratic &democratic&democratic \\
& republican &republican &republican &republican\\
\hline
&functions &functions & actor& functions\\
Topic\#59 &users & users & functions& actor\\ 
&file & file & user& user\\
\hline
& university&university&university &university\\
Topic\#56 &college &college&college&education \\ 
&education& education&education&college \\
\hline
&music& music&music&music \\
Topic\#38 &song& song& song& song \\ 
& single& single& single& single \\
\hline
\end{tabular}
\end{table}
\section{ADMM-based Networked SVI}
Now, after conclusive results about distributed `fully-conected' SVI algorithm, we move on to a network of nodes having independent learners residing at each node. The only difference in problem formulation, as we will see is in the equality constraints. We use the network formulation as given in \cite{zhang2015secure}.

\begin{figure}[!t]
\centering
\includegraphics[scale=0.75]{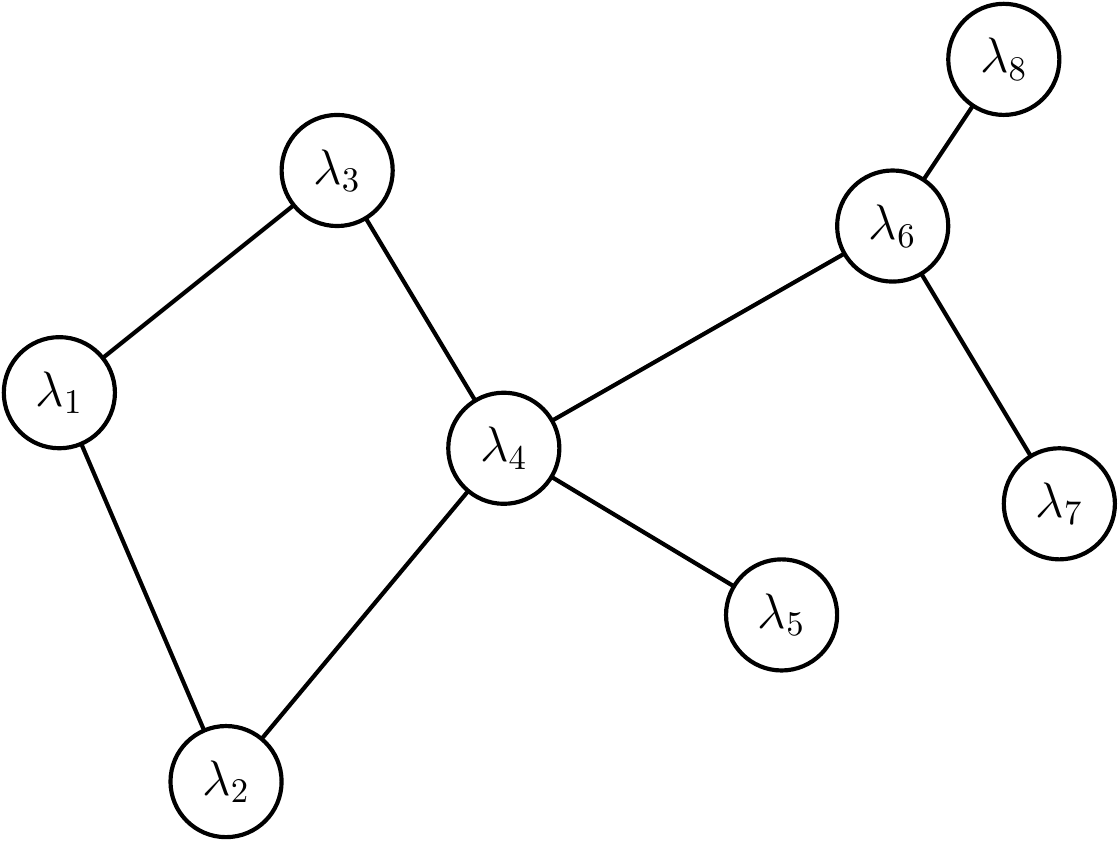}
\caption{Example of an undirected graph $\mathcal{G}(8,8)$ having $8$ nodes and $8$ edges. Here, node $4$ has the highest number of neighbors $B_4 = \{2,3,5,6\}$.\label{fig:network}}
\end{figure}

\begin{quote}\small
The network is modeled by an undirected graph denoted by $\mathcal{G}(\mathcal{K}, \mathcal{E})$ with $\mathcal{K}:=\{1,\dots,K\}$ representing the set of nodes, and $\mathcal{E}$ representing the set of links between nodes. Node $k \in \mathcal{K}$ only communicates with his neighboring nodes $B_k \subseteq \mathcal{K}$. Note that without loss of generality, graph $\mathcal{G}$ is assumed to be connected. The network can contain cycles. An example of such a network is is shown in Figure \ref{fig:network}.

At every node $k \in \mathcal{K}$, a set of observations $\mathcal{D}_k :=\{x_{kn}:n = 1, \dots, N_k\}$ of size $N_k$ is available, where $x_{kn}$ denotes the $n$-th observation for the $k$-th node. Though not explicitly expressed, each $x_{kn}$ can be a collection of multiple random variables. The vector of global hidden variables for node $k$ is $\beta_k$; its $N_k$ local hidden variables are $z_k = z_{k,1:N}$, each of which is a collection of $J$ variables $z_{kn} = z_{kn,1:J}$; the vector of fixed parameters is $\alpha_k$.
\end{quote} 
\begin{figure}[!t]
\centering
\includegraphics[scale=0.75]{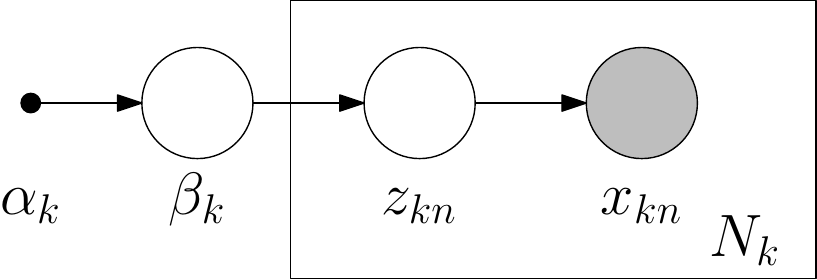}
\caption{A graphical probabilistic model for each node $k$ in the graph $\mathcal{G}$}
\end{figure}

With the graph formulation given above, we pose the distributed SVI problem for a network of learners given as:
\begin{align}
\label{eq:opt}
\begin{split}
\min_{\{\lambda_k\}} ~~~ &\sum_{k=1}^K g_k(\lambda_k)\\
\text{s.t.} ~~~ &\lambda_k = \lambda_l, \forall k\in \mathcal{K},~l\in B_k
\end{split}
\end{align}
with $\lambda_1,\cdots,\lambda_K$ as variables. Here, $g_k:\Gamma_k \to \mathbb{R}_+$ is a non-linear function of $\lambda_k$ re-written here:
\begin{align*}
g_k(\lambda_k) := -\mathbb{E}_{\phi(\lambda_k)}[\eta_g(x,z)] ^{\top} \nabla_{\lambda_k} a_g(\lambda_k) + \lambda_k^{\top}\nabla_{\lambda_k} a_g(\lambda_k)\\ - a_g(\lambda_k) + \text{const.}
\end{align*}
Optimization problem (\ref{eq:opt}) is equivalent to the following,
\begin{align}
\label{eq:opt_eq}
\begin{split}
\min_{\{\lambda_k\}} ~~~ &\sum_{k=1}^K g_k(\lambda_k)\\
\text{s.t.} ~~~ &\lambda_k = \zeta_{kl}, ~~\zeta_{kl} = \lambda_l, ~~\forall k\in \mathcal{K},~l\in B_k
\end{split}
\end{align}
where $\zeta_{kl}$ are redundant variables that will facilitate the decoupling of variable $\lambda_k$ at node $k$ from its neighboring nodes $l\in B_k$. This problem will be solved using its dual. We denote the Lagrange multipliers by $y_{kl1}$ ($y_{kl2}$) for the constraints $\lambda_k = \zeta_{kl}$ ($\zeta_{kl} = \lambda_l$). We observe that for each $k$ we have $2|B_k|$ equality constraints. The augmented Lagrangian with a quadratic penalty is:
\begin{align}
\label{eq:aug_lag_1}
\begin{split}
L_c\left(\{\lambda_k\},\{\zeta_{kl}\},\{y_{klj}\}\right) \!= \sum_{k=1}^K \!\Bigg( g_k(\lambda_k) + \sum_{l\in B_k} \! \!\Big( y_{kl1}^\top (\lambda_k - \zeta_{kl})\\ + y_{kl2}^\top (\zeta_{kl} - \lambda_l) + \frac{c}{2}\big(||\lambda_k-\zeta_{kl}||^2 + ||\zeta_{kl}- \lambda_l||^2 \big)\Big)\Bigg),
\end{split}
\end{align}
The augmented Lagrangian can be iteratively minimized with respect to each variable by keeping others constant, which gives us a set of minimization updates for each variable summarized in the following Proposition.
\subsection{Proposition 1}
The distributed iterations solving (\ref{eq:opt_eq}) are as follows:
\begin{align}
\label{eq:admm_it1}{\lambda}_k^{t+1} &=  \arg\min_{\lambda_k} \left\lbrace \begin{aligned} g_k(\lambda_k) + \lambda_k^\top \sum_{l\in B_k} \left(  y_{kl1}^t  - y_{lk2}^t \right)\\ + \frac{c}{2}\sum_{l\in B_k} \left( ||\lambda_k-\zeta_{kl}^t||^2 + ||\zeta_{lk}^t-\lambda_k||^2\right) \end{aligned}\right.\\
\label{eq:admm_it2}
\zeta_{kl}^{t+1} &= \arg\min_{\zeta_{kl}} \left\lbrace \begin{aligned}
- \zeta_{kl}^\top {y_{kl1}^t} + \zeta_{kl}^\top {y_{kl2}^t} + \frac{c}{2}\left(||\lambda_k^{t+1}-\zeta_{kl}||^2\right)\\
 + \frac{c}{2}\left(||\zeta_{kl}-\lambda_l^{t+1}||^2\right)
\end{aligned}\right.\\
\label{eq:admm_it3}
y_{kl1}^{t+1} &= y_{kl1}^t + c (\lambda_k^{t+1}-\zeta_{kl}^{t+1}),\quad\forall k\in \mathcal{K},~l\in B_k\\
\label{eq:admm_it4}
y_{kl2}^{t+1} &= y_{kl2}^t + c (\zeta_{kl}^{t+1} -\lambda_l^{t+1}),\quad\forall k\in \mathcal{K},~l\in B_k
\end{align}
and correspond to the standard ADMM solver discussed in \cite{boyd2011distributed}.\\

\subsubsection*{Proof} The first task is to cast the problem (\ref{eq:opt_eq}) into standard ADMM problem form in \cite{boyd2011distributed}. The network description adopted here is similar to the one used in \cite{forero2010consensus} and thus we use it to establish equivalence with standard ADMM \cite{boyd2011distributed}. Thereby, the remaining form of the minimization updates is directly derived from the augmented Lagrangian given in (\ref{eq:aug_lag_1}). The $\lambda$-minimization update (\ref{eq:admm_it1}) is derived by eliminating the terms that do not affect the minimization in augmented Lagrangian:
\begin{align*}
\lambda_k^{t+1} &= \arg\min_{\lambda_k} L_c(\lambda_k,\{\zeta_{kl}^t\},\{y_{kl1}^t\},\{y_{kl2}^t\}),\\
&=  \arg\min_{\lambda_k} \left\lbrace \begin{aligned}
g_k(\lambda_k) + \sum_{l\in B_k}\! \left(  \lambda_k^\top y_{kl1}^t+ \frac{c}{2}||\lambda_k-\zeta_{kl}^t||^2\right)\\
+\sum_{s\in B_k} \left( - \lambda_k ^\top y^t_{sk2}+ \frac{c}{2}||\zeta_{sk}^t-\lambda_k||^2\right)
\end{aligned}\right.,
\end{align*}
which upon merging the two summations reduces to (\ref{eq:admm_it1}). Similarly, the $\zeta$-minimization update (\ref{eq:admm_it2}) comes directly from,
\begin{align*}
\zeta_{kl}^{t+1} = \arg\min_{\zeta_{kl}} L_c(\{\lambda_k^{t+1}\},\zeta_{kl},\{y_{kl1}^t\},\{y_{kl2}^t\}).
\end{align*}
Equations (\ref{eq:admm_it3})--(\ref{eq:admm_it4}) are the dual variable updates (cf. \cite{forero2010consensus}).
\hfill$\blacksquare$

Next we reduce the iteration equations to a simpler form. Here, we observe that the $\zeta_{kl}$ update has the following unique solution (by putting the derivative equal to zero and solving),
\begin{align}
\label{eq:zeta_kl_}
\zeta_{kl}^{t+1} = \frac{1}{2c}(y_{kl1}^t - y_{kl2}^t) + \frac{1}{2}(\lambda_k^{t+1}+\lambda_l^{t+1})
\end{align}

Putting (\ref{eq:zeta_kl}) in (\ref{eq:admm_it3})--(\ref{eq:admm_it4}) gives,
\begin{align}
\label{eq:y_kl_1}y_{kl1}^{t+1} ~=&~ \frac{1}{2}(y_{kl1}^t+y_{kl2}^t) + \frac{c}{2} (\lambda_k^{t+1}-\lambda_l^{t+1}),\\
\label{eq:y_kl_2}y_{kl2}^{t+1} ~=&~ \frac{1}{2}(y_{kl1}^t+y_{kl2}^t) + \frac{c}{2} (\lambda_k^{t+1}-\lambda_l^{t+1}).
\end{align}
Now, we assume that both the Lagrange multipliers are identically initialized at every node $k$, as zero $y_{kl1}^0 = y_{kl2}^0 = 0_{m \times 1}\quad\forall k\in \mathcal{K},~l\in B_k$. This ensures that $y_{kl1}^1 = y_{kl2}^1$, and $y_{kl1}^2 = y_{kl2}^2$, and son on. We see that only one of the two multipliers per node needs to be updated at each time step. Furthermore, (\ref{eq:zeta_kl_}) simplifies to,
\begin{align}
\label{eq:zeta_kl}
\zeta_{kl}^{t+1} = \frac{1}{2}(\lambda_k^{t+1}+\lambda_l^{t+1}).
\end{align}
Finally the ADMM iterations $\forall k\in \mathcal{K}$ simplify, summarized in the following Proposition.

\subsection{Proposition 2}
Selecting $y_{k}^0 := y_{kl1}^0 = y_{kl2}^0 = 0_{m \times 1}$ as initialization $\forall k\in \mathcal{K},~l\in B_k$, the iterations (\ref{eq:admm_it1})--(\ref{eq:admm_it4}) reduce to the following,
\begin{align}
\label{eq:lam_update_final}
\lambda_k^{t+1}\! &=\arg\min_{\lambda_k}\begin{aligned}g_k(\lambda_k) +  \lambda_k ^\top y_{k}^t + c\!\sum_{l\in B_k}\!\left\lvert \left\lvert\lambda_k-\frac{1}{2}(\lambda_k^t+\lambda_l^t)\right\rvert\right\rvert ^2
\end{aligned}\\
y_{k}^{t+1}\! &= y_{k}^t + c\sum_{l\in B_k} (\lambda_k^{t+1}-\lambda_{l}^{t+1}),\quad\forall k\in \mathcal{K}
\end{align}
\subsubsection*{Proof} Substituting (\ref{eq:zeta_kl}) into the objective (\ref{eq:admm_it1}) gives the following,
\begin{multline}\label{eq:prop_2_1}
\arg\min_{\{\lambda_k\}} L_c(\{\lambda_k\},\{\lambda_k^t\},\{\zeta_{kl}^t\},\{y_{kl1}^t\},\{y_{kl2}^t\})=\\
\sum_{k=1}^K \left( g_k(\lambda_k) + \lambda_k^\top \sum_{l\in B_k} \left(  y_{kl1}^t  - y_{lk2}^t \right) + \right.\\\left. \frac{c}{2}\sum_{l\in B_k} \left( \left|\left|\lambda_k-\frac{1}{2}(\lambda_k^{t}+\lambda_l^{t})\right|\right|^2 + \left|\left|\frac{1}{2}(\lambda_l^{t}+\lambda_k^{t})-\lambda_k\right|\right|^2\right)\!\!\right).
\end{multline}
Note that $\{\lambda_k\}$ is the set of all variables of optimization and $\{\lambda_k^t\}$ denote constants known from previous iteration. All-zero initialization of the Lagrange multipliers implies that $y_{kl1}^t = -y_{lk1}^t \forall t$ [cf. from (\ref{eq:y_kl_1})--(\ref{eq:y_kl_2})], and so the first double sum in (\ref{eq:prop_2_1}) can be rewritten as:
\begin{align}
\label{eq:prop_2_2}
\sum_{k=1}^K\sum_{l\in B_k} \lambda_k^\top \left(  y_{kl1}^t  - y_{lk2}^t \right) = 2\sum_{k=1}^K \lambda_k^\top \sum_{l\in B_k}y_{kl1}^t.
\end{align}
The other two double sums in (\ref{eq:prop_2_1}) can be simplified to give,
\begin{multline}
\label{eq:prop_2_3}
\frac{c}{2}\sum_{l\in B_k} \left( \left|\left|\lambda_k-\frac{1}{2}(\lambda_k^{t}+\lambda_l^{t})\right|\right|^2 + \left|\left|\frac{1}{2}(\lambda_l^{t}+\lambda_k^{t})-\lambda_k\right|\right|^2\right)\\
=c\sum_{l\in B_k} \left( \left|\left|\lambda_k-\frac{1}{2}(\lambda_k^{t}+\lambda_l^{t})\right|\right|^2\right).
\end{multline}
By defining $y_k^t := 2\sum_{l\in B_k} y_{kl1}^t$, and substituting (\ref{eq:prop_2_2}) and (\ref{eq:prop_2_3}) into (\ref{eq:prop_2_1}), gives the final form of the augmented Lagrangian which completes the proof:
\begin{multline*}
\arg\min_{\{\lambda_k\}} L_c(\{\lambda_k\},\{\lambda_k^t\},\{y_{k}^t\}) = \sum_{k=1}^K g_k(\lambda_k) +  \sum_{k=1}^K\lambda_k ^\top y_{k}^t  \\ + c\sum_{k=1}^K\sum_{l\in B_k}\left\lvert \left\lvert\lambda_k-\frac{1}{2}(\lambda_k^t+\lambda_l^t)\right\rvert\right\rvert ^2.
\end{multline*}
\hfill$\blacksquare$

\subsection{Network solution}
Now, we present a solution to the ADMM minimization update (\ref{eq:lam_update_final}), which is a non-convex optimization problem, similar to the corresponding update in distributed SVI in section~\ref{sec:dist_svi_solution}. 
We make use of stochastic gradient descent like standard SVI algorithm for minimization of augmented Lagrangian (cf. \cite{hoffman2013stochastic}). It is known that the natural gradient of $g_k$ is given as,\[\hat{\nabla}_{\lambda_k} g_k(\lambda^t_k) = \lambda_k^t - \hat{\lambda}_k.\]The solution is presented in algorithm~\ref{algo:svi-netw-admm}.
\begin{algorithm}
\caption{ADMM-based networked SVI for $K$ players}
\label{algo:svi-netw-admm}
\begin{algorithmic}[1]
\State{Initialize $c,\lambda_1^{(0)},\lambda_2^{(0)},\dots,\lambda_K^{(0)}$}
\State{Schedule step-size $\rho_t$ routine}
\Repeat
\For{$k\in\mathcal{K}$}
\State{Sample separate data points $x_k$ for all learners}
 \State{Use $x_k$ to compute its local variational parameters,
 \[\phi = \mathbb{E}_{\lambda_k^{t}}[\eta_l(x_k^{(N)},z_{k}^{(N)})].\]}
\State{Apply ADMM $\lambda$-minimization-update by computing intermediate global parameters $\hat{\lambda}_k$ and natural gradient of augmented Lagrangian,
\[\hat{\lambda}_k = \mathbb{E}_{\phi}[\eta_g(x_k^{(N)},z_k^{(N)})],\]
\[\hat{\nabla}_{\lambda_k^t} L_c = (\lambda_k^t -\hat{\lambda}_k) - \left[\nabla^{-2} a_g(\lambda_k^t)\right]\left(y_k^t + c\sum_{l\in B_k}\left(\lambda_k^t-\lambda_l^t\right)\right).\]
}
\State{Update the global variational parameters using gradient {ascent},
\[\lambda^{t+1}_k = \lambda^{t}_k + \rho^t \left(-\hat{\nabla}_{\lambda_k^t} L_c\right).\]}
\EndFor
\State{Update all the Lagrange multipliers
\[y_{k}^{t+1} = y_{k}^t + c\sum_{l\in B_k} \left(\lambda_k^{t+1}-\lambda_{l}^{t+1}\right).\]}
\Until{forever}
\end{algorithmic}
\end{algorithm}
\subsection{Experimental results}
The issue of cross-matching the topics between two different players was solved using a correlation metric. Like discussed earlier, SVI relies on random initialization of the global parameters. So, each player initializes with different global parameters, and as they encounter observations, they update the global parameters. Since, in our setting each player has its own independent dataset, so the trajectory of converging to `true' topics is different for every player. That is why, if two completely independent learners are fed the same data, they converge to similar estimates but with different trajectories (i.e. topic $0$ for player A may truly represent the contents of topic $43$ for player B, and so on). Thus, in order to match the right topics, a correlation metric was needed. We used the Pearson correlation coefficient for this.

\begin{figure}[!t]
\centering
\subfloat[]{$\vcenter{\hbox{\includegraphics[width=0.3\linewidth]{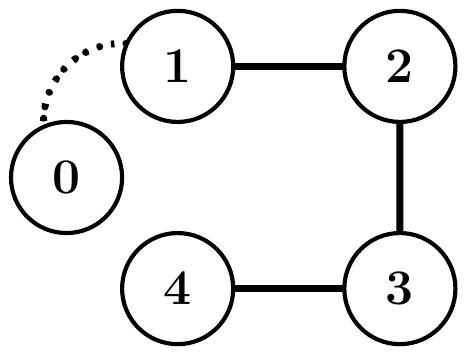}\label{fig:network_v2}}}$}
\hfill
\subfloat[]{$\vcenter{\hbox{\includegraphics[width=0.7\linewidth]{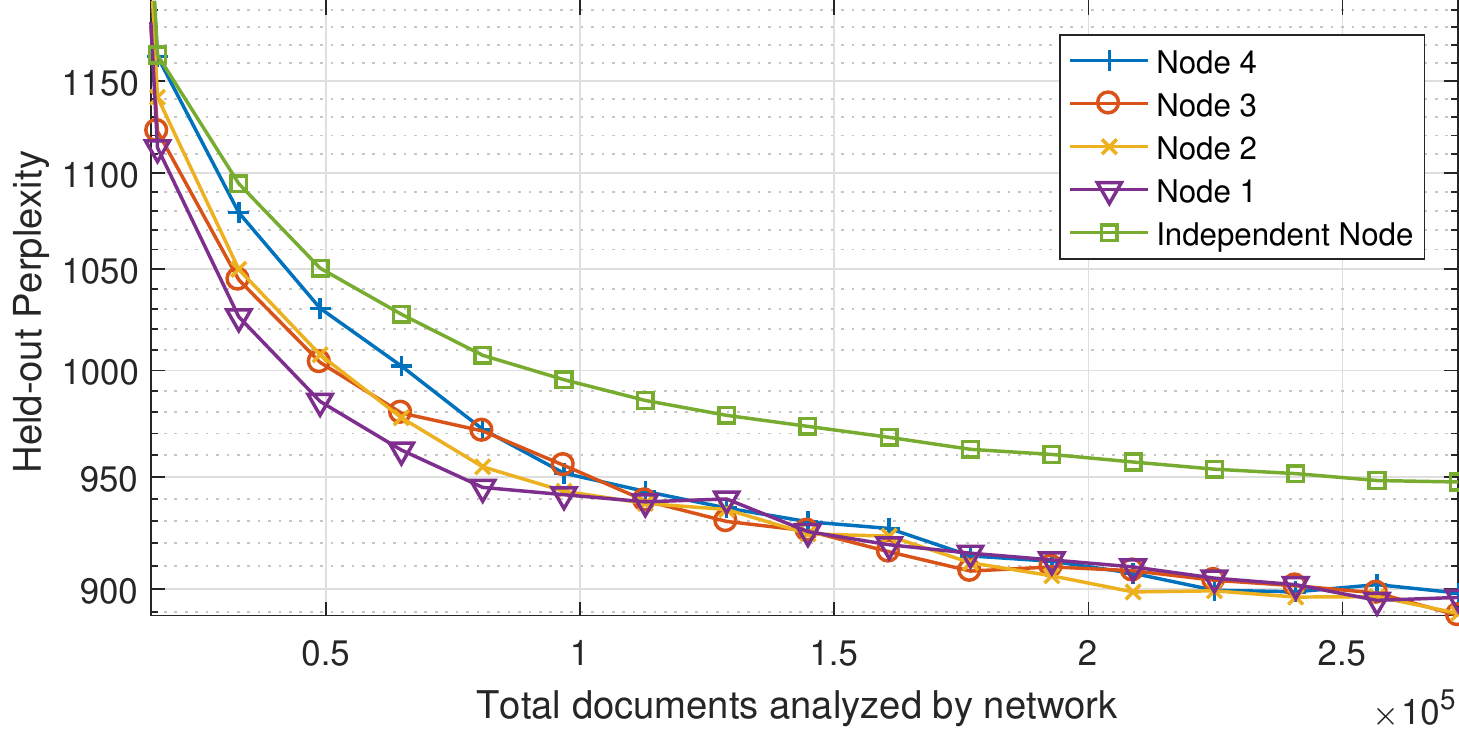}\label{fig:perp_exp7}}}$}
\caption{(a) Line-type graph network. Dotted line between two nodes indicates same dataset supply. Solid line indicates possibility of transfer learning between nodes via ADMM. (b) Perplexity trajectory for a line-type graph.}
\end{figure}

Result for an experiment that used a line-type graph is shown in Fig.~\ref{fig:perp_exp7}. The network is shown in Fig.~\ref{fig:network_v2}. In this experiment nodes $0$ and $1$ were provided with exactly same set of data (limited to a fixed $800$ documents offline available that were fed repeatedly). The nodes $2$, $3$, and $4$ were provided with an online data i.e. independent and new data points at each iteration. The purpose of this experiment was to see how the connected nodes corroborate and improve estimation accuracy (of node $1$) in contrast to accuracy of the independently running learners (i.e. node $0$). The perplexity metric trajectory shown in Fig.~\ref{fig:perp_exp7}, supports our claim that node interaction through ADMM updates certainly benefits. We achieve better accuracy in the estimate of node $1$ as compared to that of node $0$ because the learning at nodes $2$-$4$ affects that of node $1$ due to the consensus constraint.
\begin{figure}[!t]
\centering
\subfloat[]{$\vcenter{\hbox{\includegraphics[width=0.3\linewidth]{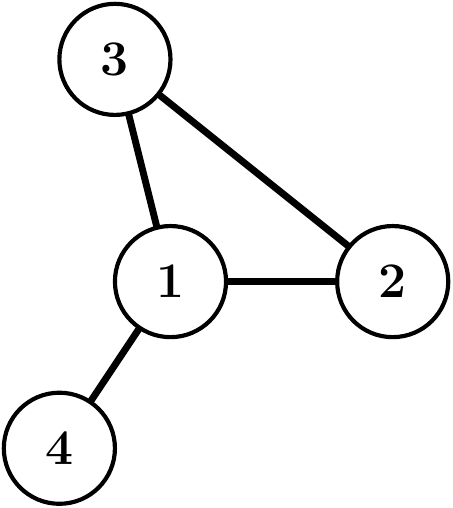}\label{fig:network_v3}}}$}
\hfill
\subfloat[]{$\vcenter{\hbox{\includegraphics[width=0.7\linewidth]{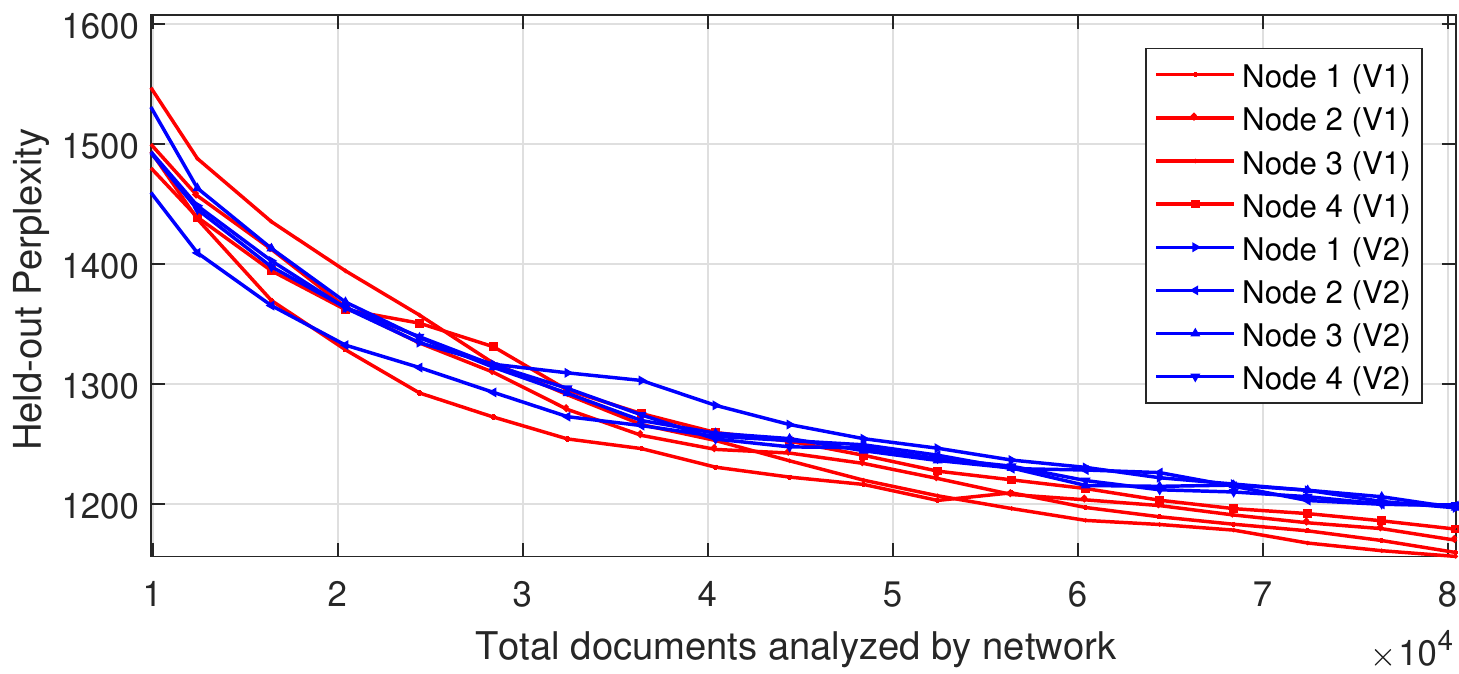}\label{fig:perp_exp8}}}$}
\caption{(a) Star-type strongly connected network. (b) Perplexity trajectory for a strongly connected network (V1) versus a weakly connected line-type network (V2). Clearly strongly connected network starts performing better after some iterations.}
\end{figure}
\begin{figure}[!t]
\centering
\includegraphics[width=\linewidth]{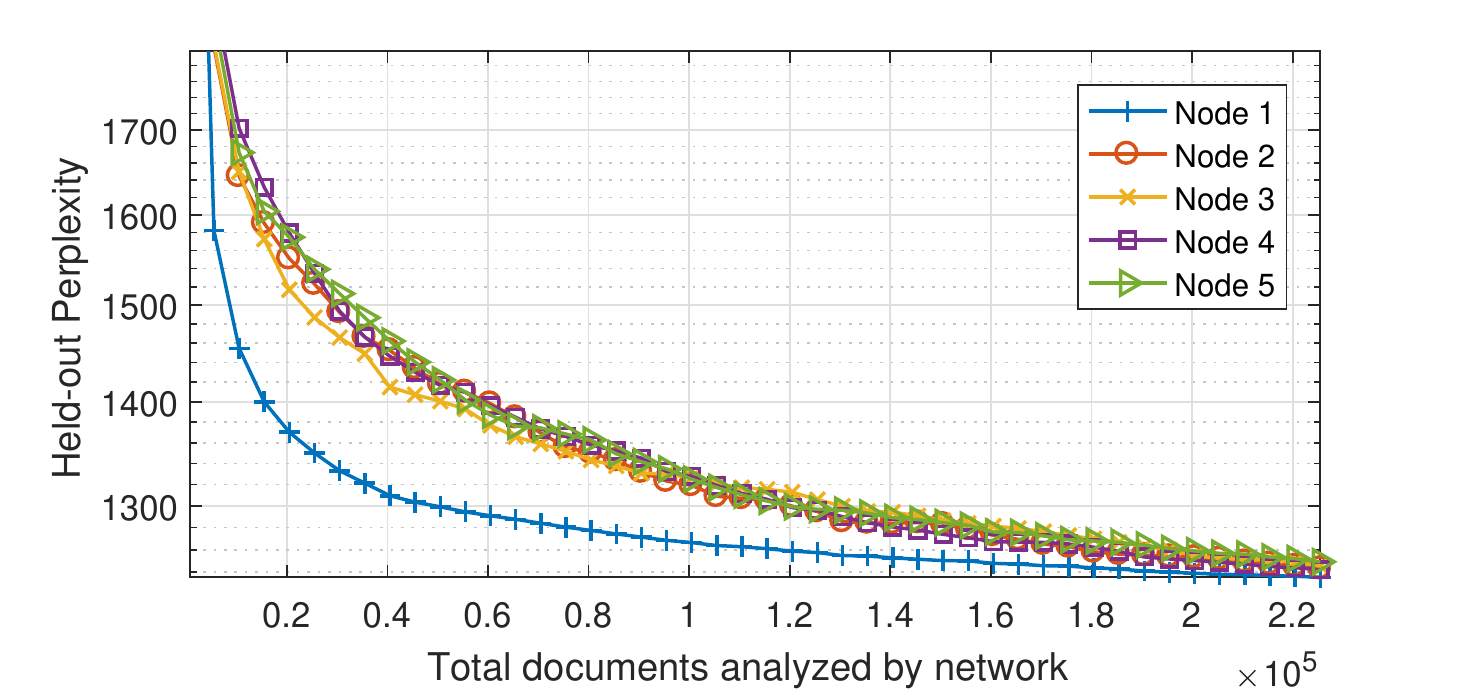}
\caption{Independent SVI with complete data versus networked SVI with partitioned data. Node 1 is an independent learner. Nodes 2-5 are connected in a network having partitioned datasets.}\label{fig:network_v4}
\end{figure}
\section{Discussion and Conclusion}
We have presented distributed ADMM-based SVI and an extension of it over a network of learners in a grpah -- an algorithm that solves separable stochastic optimization problems and merges their results to achieve optimal consensus solution. Applications of distributed learning agent systems are common in IoT framework especially when different learning systems do not want to share data among each other but still agree on partial collaboration and transfer learning. A well-trod example of latent Dirichlet allocation for probabilistic topic models is implemented to show comparative results for the centralized, distributed (thoroughly fully-connected) and networked settings. One key observation in the results of our networked SVI algorithm is that strongly connected networks exhibit substantial transfer learning benefits. This is highlighted in the comparative experiments of Fig.~\ref{fig:network_v2} and Fig.~\ref{fig:network_v3}. Another observation is that accuracy of estimation improves over time, as more and more data is analyzed. In a nutshell, results show that through collaboration without having to share private data, two or more independent model posterior learners for SVI can improve their learning capabilities. Due to the use of stochastic optimization, this algorithm is considerably fast, scalable, and accurate. Moreover, its distributed learning methodology enhances security and robustness aspects that underpin modern deep learning goals. For future, we intend to apply this to cyber-physical dynamical systems along with inspecting guarantees on the convergence properties of this algorithm.

\ifCLASSOPTIONcaptionsoff
  \newpage
\fi


\bibliographystyle{IEEEtran}
\bibliography{./ref}
\newpage
\vfill
\end{document}